# Multi-task Learning for Macromolecule Classification, Segmentation and Coarse Structural Recovery in Cryo-Tomography

Chang Liu[1], Xiangrui Zeng[1], Kaiwen Wang[1], Qiang Guo[2], and Min Xu [*1]

[1]Carnegie Mellon University, Pittsburgh, PA, USA
[2]Max Planck Institute for Biochemistry, Martinsried, Germany

**Abstract**

Cellular Electron Cryo-Tomography (CECT) is a powerful 3D imaging tool for studying the native structure and organization of macromolecules inside single cells. For systematic recognition and recovery of macromolecular structures captured by CECT, methods for several important tasks such as subtomogram classification and semantic segmentation have been developed. However, the recognition and recovery of macromolecular structures are still very difficult due to high molecular structural diversity, crowding molecular environment, and the imaging limitations of CECT. In this paper, we propose a novel multi-task 3D convolutional neural network model for simultaneous classification, segmentation, and coarse structural recovery of macromolecules of interest in subtomograms. In our model, the learned image features of one task are shared and thereby mutually reinforce the learning of other tasks. Evaluated on realistically simulated and experimental CECT data, our multi-task learning model outperformed all single-task learning methods for classification and segmentation. In addition, we demonstrate that our model can generalize to discover, segment and recover novel structures that do not exist in the training data.

## 1 Introduction

The cell is the basic unit of living organisms. Most cellular processes are governed by macromolecules. To fully understand such processes, it is necessary to precisely know the structure and spatial organization of all macromolecules inside single cells. Such information has been extremely difficult to obtain due to limitations in data acquisition. Recently, Cellular Electron Cryo-Tomography (CECT) has emerged as a dominating 3D imaging technique that captures cellular structure at sub-molecular resolution and in close-to-native state, providing systematic 3D visualization of close-to-native state macromolecular structures in unprecedented resolution and fatefulness. However, systematic recovery of macromolecule structures is very challenging due to the imaging limitation of CECT and structural complexity of macromolecules. In particular, the images are captured at a very low signal-to-noise ratio (SNR), making it hard to identify macromolecules via simple inspection. Because of the limited imaging tilt angle range at data acquisition stage, CECT 3D images (aka tomograms) generally suffer from missing wedge effect (i.e. missing values), resulting in anisotropic resolution. Also, macromolecules are structurally highly diverse. They are often densely distributed inside the tomograms while dynamically interacting with each other, therefore introducing more complex and heterogeneous structures.

Earlier studies of macromolecule localization inside CECT image relied on template matching[3], comparing an experimental macromolecule to a known structural template. For the unbiased detection, classification, and high-resolution structure recovery of all heterogeneous complexes, reference-free method are needed. Several unsupervised approaches have been developed on macromolecular structure recovery. The pipeline consists of three main steps: first, particle-picking methods [17] are applied for the localization of potential macromolecules and then subtomograms[1] are extracted from a tomogram based on those locations. Second, due to the crowded cellular environment of tomogram, extracted subtomograms may contain not only the target macromolecule but also fragments of neighboring structures. Previously, model based clustering method [19] was introduced for automatic segmentation of the target complex. However, those manually heuristic rules were not reasonable enough to deal with the inherent complexity of CECT data. For instance, the method assumes that all structural regions are spherical-like while rod or plane like structures are not being taken into account. Third, reference-free subtomogram alignment, classification and averaging methods [e.g. 5, 2, 18, 20, 8, 22] subdivide all macromolecules into several homogeneous groups, and the averaging of the subtomograms of same macromolecular structure will improve the resolution of recovered structure, achieving structure recovery from CECT data.

---

*Corresponding author email:mxu1@cs.cmu.edu.

Despite that the promising classification and structure recovery results have been shown, existing unsupervised approaches still suffer from limited scalability and discrimination ability due to intensive computations. Recently, deep learning based methods have been analyzed to assist systematic structure recovery pipeline while maintaining great scalability and accuracy. In particular, Liu et al [15] have proposed a 3DCNN based segmentation network, named SSN3D-ED which is capable of masking out neighboring structures inside subtomograms while maintaining generalization ability to segment different types of macromolecules. Also, large-scale 3D subtomogram classification models [21, 7] such as DSRF3D-v2 have achieved solid results on subtomograms which contain no neighboring structures. Then reference-free structural recovery approaches can be applied to those classified homogeneous group of subtomograms.

Generally, deep learning based subtomogram segmentation and classification are performed separately in a cascade manner, and therefore require training and inference on different models, complicating the automatic structure recovery pipeline while taking more computational resource. Since low-level image features in the 3DCNN models[21, 15] can be shared by both tasks, it is a natural and intuitive idea to build up such a multi-task model to complete those two steps end-to-end in one shot while improving the performance via inductive transfer[6] compared to single-task learning.

In this paper, three CECT analysis tasks have been explored: 1) Identify the structural class of the macromolecule contained in a subtomogram if it is a known structure class in training data; 2) semantically segment the target macromolecule out of neighbor structures in subtomograms; 3) an auxiliary task of coarsely recovering the density map of macromolecular structure for the assistance of two main tasks and proof-of-principle visualization. We proposed a novel 3DCNN multi-task learning model named Deep Subtomogram Multi-task Network (DSM-Net, **Fig**.1) for implementing those three tasks.

Tests on both realistically simulated and experimental dataset, our multi-task model significantly outperformed the single-task models that separately doing classification and segmentation. In addition, we show that our model has certain generalization ability to classify, segment and coarsely recover the new structures that do not exist in the training data.

## 2 Method

DSM-Net is an end-to-end, unified network which consists of a backbone network for computing convolutional feature maps, and three parallel subnetworks for subtomogram segmentation, classification and coarse structure recovery.

We use the low level image features shared by the 3DCNN models[21, 15] to build up a multi-task model that performs semantic segmentation and subtomogram classification jointly. Further, we redefine the structure recovery process as a image reconstruction problem, and extend the model by adding a small overhead on backbone to jointly learn the coarse recovered structures (represented by a density map) of target macromolecule.

### 2.1 Residual based Backbone Network and Classification Subnet

Considering the scenario that input subtomograms contain not only target complexes but also neighboring structures, previous 3D VGG based subtomogram classification model such as DSRF3D-v2 [7] failed to converge during the training process. To increase the optimization ability, 3D residual block plus stride-2 convolutional layer design [12] is introduced to the backbone network. In particular, input is connected to 3D conv followed by stride-2 maxpooling layer. Then two consecutive 3D residual block plus a stride-2 3D conv layer are applied. Finally, it is followed by a 3D residual block plus 3D average pooling layer. The number of channels of three residual block is (32, 64, 128) respectively.

---

[1] A *subtomogram* is a small cubic subvolume of a tomogram generally contains a single macromolecule extracted from a tomogram.

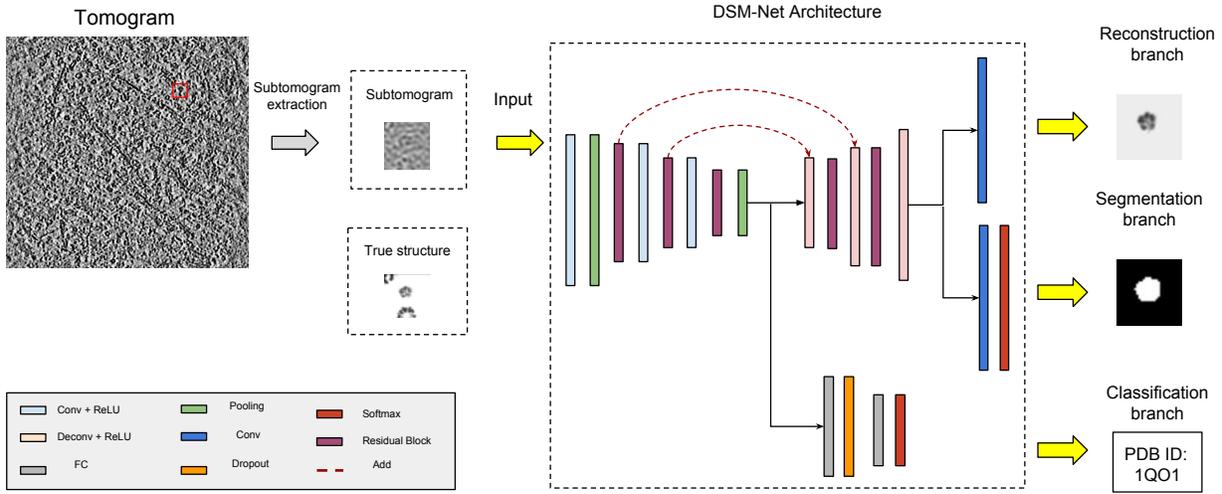

Figure 1: A conceptual diagram of DSM-Net architecture.

For classification subnet, it predicts the probability of input subtomogram categories (C = 22 classes). The backbone network is followed by a fully connected layer with 1024 hidden units at the dropout rate of 70%. Equipped with a softmax activation layer, the subnet outputs C-unit vector and predicts query subtomogram to the macromolecular class that returns the highest probability.

Note that all following convolution and deconvolution layers are followed by a ReLU activation and their kernel size is $3 \times 3 \times 3$ except the final convolution layers of segmentation and structure recovery subnet.

## 2.2 Segmentation Subnet

The backbone network and segmentation subnet follow the encoder-decoder architecture adopted from previous model SSN3D-ED which is a 3D variant of fully convolutional network. Except 3D residual block is taken for computing feature maps at backbone network, 3D deconvolution replaces interpolation-based upsampling layers in segmentation subnet compared to SSN3D-ED. In our experiments, we found that this in-network upsampling filters outperformed simple bilinear upsampling ones for learning dense pixel prediction.

Specifically, the backbone is followed by two consecutive 3D stride-2 deconv plus a 3D residual block. Then, one more 3D stride-2 deconv is attached and followed by a $1 \times 1 \times 1$ 3D conv layer with the channel size of 2 (target region and background). After applied softmax layer, predictions is voxel-level classification. In addition, skip connections from the lower layers of residual block are added to the correspondingly higher deconvolution layers, in order to integrate coarse information with fine-grained information for solving local ambiguities.

## 2.3 Structure Recovery Subnet

After subtomograms are well segmented and classified, structure recovery is usually achieved through unsupervised alignment and the averaging of homogeneous group of complexes to obtain a high-resolution density map. For improving scalability while keeping recovery accuracy, we redefine the structure recovery as a supervised image reconstruction problem. Given $x$ is a measurement of 3D CECT image, $G$ is the unsupervised model whose response is regard as the ground truth of structure recovery and $H$ is our supervised DSM-Net model, the objective function of image reconstruction can be written as:

$$L_{rec} = \frac{1}{N} \sum_x ||G(x) - H(x)||_2 \qquad (1)$$

where N is the number of training examples.

Structural recovery and segmentation of the macromolecule of interest are highly related tasks. Specifically, the structural recovery determines the electron density of the macromolecule of interest at each voxel location which directly correlates to subtomogram image intensity, whereas the semantic segmentation is a binary decision on whether a voxel is occupied by the macromolecule of interest by taking into account of the combination of subtomogram image intensity. With respect to distinction, segmentation task generally segments only a rough region of target complex while recovery task focuses even more on the fine-grained contour and inner cellular structure as well. One example have shown this distinction: a hole in a macromolecule will show a low electron density on the density map, but still be categorized as a part of the segment (see first column of **Fig**. 2 c and d for example).

Shared parameters with segmentation task except the final convolution layer, structure recovery subnet simply add a 3D conv with channel = 1 to predict a reconstructed image while keeping model complexity. In this paper, we regard the structure recovery only as an auxiliary task which might improve the generalization ability of two main tasks and provide 3D visualization clues for discovering new structures. Further experiments which quantitatively evaluate this task and task-specified network architecture will be explored in future.

## 3 Experiments

We evaluated our proposed DSM-Net on realistically simulated and experimental dataset and compared the result to previous single-task models: DSRF3D-v2 and SSN3D-ED. Additionally, We made comprehensive ablations experiments in task-specified single models extracted from DSN-Net respectively. Further, sensitivity tests on the combination of weighted loss ratio of three tasks have been conducted.

### 3.1 Dataset

**Simulated subtomograms from known structures** For a persuasive assessment of the approach, we generated realistically simulated tomograms with known structures of macromolecular complexes (class of macromolecules of the same structure) by simulating the actual tomographic image reconstruction process as previously described[17]. The limitation of CECT data such as missing wedge, image noise, electron optical factors, including the Modulation Transfer Function (MTF) and the Contrast Transfer Function (CTF), were properly included.

Specifically, 22 distinct macromolecular complexes (Tab.1) are chosen from the Protein Databank (PDB) [4] for experiments. Each simulated tomograms of $600 \times 600 \times 300$ voxels contains 10000 randomly distributed macromolecules with a tilt angle range $\pm 60°$. Given the true position of these macromolecules inside tomograms, we extracted the subvolumes of $40^3$ voxels centering on these positions as input to our model. Removing those subtomograms outside the boundary of tomograms, we finally collected 3205 simulated subtomograms of 22 structural classes for each dataset. Datasets A, B have SNR of 0.06, and 0.01 respectively.

**Experimental tomograms** A ribosome dataset of 859 subtomograms was extracted from a tomogram of primary rat neuron culture [11]. The tomogram was captured from tilt angle $-50°$ to $+70°$. It was then binned twice to a voxel size of 1.368 nm. Subtomograms of size $40^3$ were extracted from the tomogram using Difference of Gaussian particle picking method [17] and coarsely filtered by a convolutional autoencoder [23]. Template search was applied to select the top 1000 subtomograms with highest structural correlation with the ribosome template. We manually inspected the 1000 subtomograms, and filtered out 141 of them which contained obvious non-ribosome structure such as fiducial.

Additionally, a dataset consisting of 386 single capped proteasome subtomograms is extracted from a tomogram of rat neuron with expression of poly-GA aggregate [11]. All subtomorgams were two times binned to size $40^3$ (voxel size: 1.368 nm). The tilt angle range was $-50°$ to $+70°$.

| PDB ID | Macromolecular Complex |
|---|---|
| 1A1S | Ornithine carbamoyltransferase |
| 1BXR | Carbamoyl phosphate synthetase |
| 1EQR | Aspartyl-TRNA synthetase |
| 1F1B | E. coli asparate transcarbamoylase P268A |
| 1FNT | Yeast 20S proteasome with activator PA26 |
| 1GYT | E. coli Aminopeptidase A |
| 1KPB | GroEL |
| 1LB3 | Mouse L chain ferritin |
| 1QO1 | Rotary Motor in ATP Synthase |
| 1VPX | Transaldolase |
| 1VRG | Propionyl-CoA carboxylase |
| 1W6T | Octameric Enolase |
| 1YG6 | ClpP |
| 2AWB | Bacterial ribosome |
| 2BO9 | Human carboxypeptidase A4 |
| 2BYU | M.tuberculosis Acr1(Hsp 16.3) |
| 2GHO | Thermus aquaticus RNA polymerase |
| 2GLS | Glutamine Synthetase |
| 2H12 | Acetobacter aceti citrate synthase |
| 2IDB | 3-octaprenyl-4-hydroxybenzoate decarboxylase |
| 2REC | RECA hexamer |
| 3DY4 | Yeast 20S proteasome |

Table 1: The experimental macromolecular complexes used for tomogram simulation and semantic segmentation.

To prevent class imbalance problem, we randomly select 400 ribosome subtomograms out of 859 filtered subtomograms. Overall, 400 ribosome and 386 single capped proteasome subtomograms are combined and shuffled, named Dataset C. The segmentation and density map ground truth were prepared by aligning the corresponding structural template (PDB ID: 5T2C and 5MPA).

### 3.2 Implementation Detail

All models were trained and tested on Keras[9] with Tensorflow[1] as the back-end. The EMAN2 library [10] is used for simulating tomograms. The experiments were performed on a computer with three Nvidia GTX 1080 GPUs, one Intel Core i7-6800K CPU and 128GB memory.

Adam[13] was used to optimize the parameters. We set the learning rate to $10^{-3}$ with the batch size of 64 and exponential decay rates to $\beta1 = 0.9$ and $\beta2 = 0.99$. Categorical cross-entropy was used as the loss function for semantic segmentation and classification tasks, while mean squared error for structure recovery. All models were trained for no more than 100 epochs while the early stopping criterion was applied if the validation dataset did not improve for 15 consecutive epochs.

### 3.3 Results

Dataset A, B, C were randomly split into training, validation, and testing set at the ratio of 0.1 respectively. The performance result on simulated Dataset A is reported in Table 2. First, it shows that the separately trained subnetworks of DSM-Net outperformed previous methods such as DSRF3D-v2 and SSN3D-ED. Res-backbone for classification subnet enables the optimization on subtomograms that even contains neighboring structures while VGG-backbone failed to converge. Besides, mIoU from segmentation subnet has hugely increased by 2.39% compared to SSN3D-ED. Second, different combinations of the subnetworks of DSM-Net have been evaluated. It illustrates that all multi-task models outperformed single-task ones while DSM-Net maintained the best mIoU and object accuracy. It indicates that structure recovery branch as auxiliary task assisted the optimization of two main tasks with adding minor computational cost. Specifically, segmentation result classification accuracy reaches 93.75%, which is a substantial enhancement compared other models.

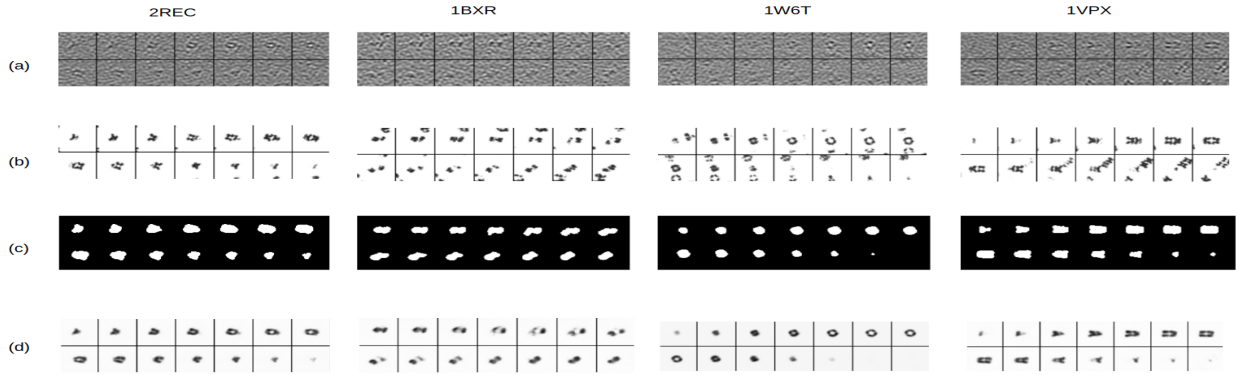

Figure 2: Examples of key 2D slices cut from input subtomograms and corresponding output segmentation predictions and regressed density maps: (a) Input subtomograms (b) True structures used to produce simulated subtomograms (c) Segmented regions of interest (d) Recovered structure of macromolecules of interest represented as density map.

Table 2: Evaluation performance on simulated Dataset A with SNR = 0.06. To clarify, classification, segmentation and structure recovery subnet are individually or dually extracted from DSM-Net, abbreviated as Cls, Seg and Rec Subnet below. '-' indicates incompatible tasks while '*' indicates the model fails to converge.

| Method | Backbone | Segmentation Pix acc | mIoU | Classification Obj acc |
|---|---|---|---|---|
| DSRF3D-v2[7] | 3D VGG-8 | - | - | * |
| Cls Subnet | 3D ResNet-9 | - | - | 76.24 |
| SSN3D-ED [15] | 3D VGG-8 | 98.99 | 84.68 | - |
| Seg Subnet | 3D ResNet-9 | 99.03 | 87.07 | - |
| Cls+Rec Subnet | 3D ResNet-9 | - | - | 81.87 |
| Cls+Seg Subnet | 3D ResNet-9 | **99.22** | 88.70 | 84.37 |
| DSM-Net | 3D ResNet-9 | 99.21 | **89.00** | **93.75** |

DSM-Net outputs are visualized in Figure 2. It is shown that neighboring structures in subtograms have been masked out (Figure 2 (c)), and the structure of target macromolecules is coarsely recovered even for some inner features such as hollow (Figure 2 (d)).

In addition, challenging Dataset B and C with SNR = 0.01 have also been assessed in Table 3. DSM-Net have shown promising results on simulated Dataset B with mIoU, and accuracy increased by 1.7% and 9.37% respectively. However, DSM-Net returns similar results on experimental dataset with respect to single-segmentation model. It is probably due to the limited number of categories in dataset C (ribosome and proteasome only) and therefore has minor influence in learning further discriminative features in shared backbone network. On the other hand, multi-task scheme solved the convergence problem on experimental subtomogramsm, achieving zero classification error.

**Ablation Experiments** Since overall loss function is a linear combination of three tasks, different pairs of loss ratio are needed to be analyzed. 2-task learning models revised from DSM-Net provides a rough suggestion on selecting such weighting coefficient of each single-task loss. From Table 4, we find that results are insensitive to those values. Finally, 1: 10: 1 for classification, segmentation and structure recovery respectively is adopted as weighting coefficient for training DSM-Net to keep the order of magnitude for three tasks at the same level.

Table 3: Evaluation performance on challenging simulated Dataset B and experimental Dataset C with SNR = 0.01.

| Method | Segmentation | | Classification | Dataset |
| --- | --- | --- | --- | --- |
| | Pix acc | mIoU | Obj acc | |
| Cls Subnet | - | - | 43.75 | |
| Seg Subnet | 98.37 | 79.70 | - | Simulated |
| DSM-Net | **98.59** | **81.40** | **53.12** | |
| Cls Subnet | - | - | * | |
| Seg Subnet | **96.01** | 60.38 | - | Experimental |
| DSM-Net | 95.80 | **61.26** | **100.00** | |

Table 4: Varying the loss ratio of given models on Dataset A.

| Method | Segmentation | | Classification |
| --- | --- | --- | --- |
| | Pix acc | mIoU | Obj acc |
| Cls+Rec 1:1 | - | - | 81.56 |
| Cls+Rec 1:10 | - | - | 81.87 |
| Cls+Seg 1:10 | 99.22 | **88.70** | **84.37** |
| Cls+Seg 1:100 | **99.23** | 88.63 | 82.81 |

## 3.4 Discussion and Insight on Model Design

The DSM-Net has an encoder-decoder architecture. Learning through the classification module of DSM-Net encourages the encoding of subtomograms to be more discriminative between different structural classes. When the encoding is fed into the segmentation decoder module, segmentation accuracy can be improved. Conversely, learning in the segmentation decoder module provides supervision on the region of interest (ROI) of macromolecule and filters out neighbor structures. The structural outline learned by segmentation decoder module contains the structural class information. The filtration of neighbor structures helps the network to focus on the macromolecular structure of interest and reduce the bias introduced by neighbor structures. The above two factors can significantly improve classification accuracy. Since image reconstruction is highly correlated to segmentation (Explained in Sec 2.3) but involving more fine-grained structural information, successful recovery of structure will make the shared upsampling module learn even detailed contour of macromolecule, and therefore improve the segmentation result.

## 3.5 Identification and recovery of unseen structure

As described in the introduction section, unsupervised reference-free structure recovery method involves clustering, alignment, and the averaging of homogeneous subtomograms, which is computationally intensive. Although 3DCNN classification model proposed by Xu et al [21] accelerates the subtomograms subdivision step, a more scalable model is needed for structure recovery to reduce computational cost.

By contrast, our proposed DSM-Net serves as an end-to-end model that can simultaneously segment, classify, and coarsely recover unseen macromolecule structures that do not exist in the training data. To test the generalization ability of DSM-Net, we modified the classification subnet of DSM-Net to output 21 units and trained DSM-Net on Dataset $A$ (containing 22 classes), excluding all the GroEL subtomograms (PDB ID: 1KP8).

For discovering this new structure, we inferred all training, and testing data via DSM-Net and outputted 1024 hidden units from the fully connected layer of classification subnet. Those hidden units interpreted as structural features are invariant to missing wedge effect and rigid transformation. After that, k-means clustering with k = 22 was performed on all nonlinear transformed data, and we picked out the cluster enriched with 1KP8 subtomograms. For visual assessment of this discovery of 1KP8, we embedded all clusters into an $\mathbb{R}^2$ space using the T-SNE algorithm [16]. According to **Fig**.3(g), it is obvious that 1KP8 subtomograms are mostly located inside a single cluster even though this structure did not exist in the training data. Additionally, the structure recovery subnet of DSM-Net can output a coarse new structure of 1KP8 (**Fig**.3(b)). The performance of structural recovery is measured using Fourier Shell Correlation [14] between true and recovered structures of 1KP8 for the representation of structural discrepancy which is reported to have a decent score 5.0.

To emphasize, 1KP8 is largely distinct from other structures in training set, and therefore the successful discovery, and sructural recovery of this structure can strongly support the generalization ability of our multi-task model.

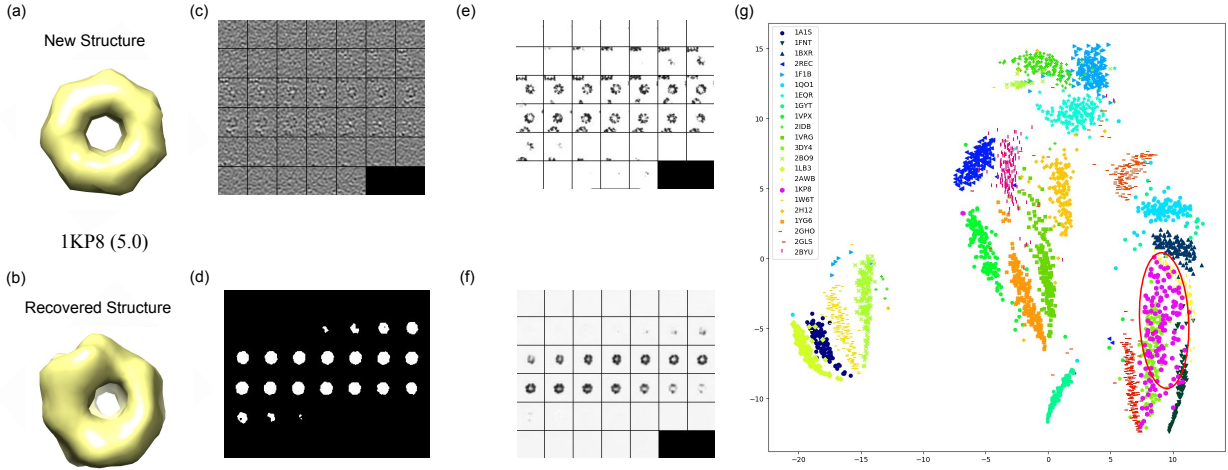

Figure 3: The outputs of DSM-Net for the new structure GroEL (PDB ID: 1KP8) with respect to training data. (a) The isosurfaces of true structure (b) The isosurfaces of recovered structure (c) Subtomogram of GroEL (d) Segmented region of GroEL (e) True structure of subtomogram (f) Recovered structure of subtomogram (g) The visualization of subtomograms in dataset $A$ embedded to $\mathbb{R}^2$ using T-SNE. The region enriched with GroEL subtomograms is highlighted using red circle.

## 4  Conclusion

In this paper, we present a novel multi-task 3D convolutional neural network (DSM-Net) for jointly performing classification, semantic segmentation, and coarse structural recovery of macromolecules in subtomograms. To our knowledge, this work is the first application of deep multi-task learning for CECT analysis. Evaluated on simulated and experimental dataset with different noise level, DSM-Net markedly outperforms two baseline single-task networks. In addition, we redefine structure recovery as a supervised image reconstruction problem, which serves as an auxiliary task for assisting two main tasks. The output of auxiliary task provides potential clues for discovering new structures. Further, we demonstrate that our model has certain generalization ability to classify and recover the structures that do not exist in training data. Our work serves as an important step towards systematic structural identification and recovery of macromolecules captured by CECT. For future works, structure recovery task will be quantitatively analyze, and improved. Also, improving performance in low SNR subtomograms and identifying macromolecule spatial interactions (such as interaction with cell membrane) though semantic segmentation are other important issues to be explored.

# 5 Acknowledgements

This work was supported in part by U.S. National Institutes of Health (NIH) grant P41 GM103712. MX acknowledges support from Samuel and Emma Winters Foundation.